\documentclass[]{spie}  

\usepackage{amsmath,amsfonts,amssymb}
\usepackage{algorithm}
\usepackage{algpseudocode}
\usepackage{authblk}
\usepackage{multirow}
\usepackage{float}
\usepackage{graphicx}
\usepackage{tcolorbox}
\usepackage{tabularx}
\usepackage{xcolor}
\tcbset{tab2/.style={colback=red!5!white,colframe=red!50!black,colbacktitle=red!40!white,
coltitle=black,center title}}
\usepackage{colortbl}
\usepackage[colorlinks=true, allcolors=blue]{hyperref}
\usepackage{multirow}

\title{	Deep neural network goes lighter: A case study of deep compression techniques on automatic RF modulation recognition for Beyond 5G networks}
\author{
  $\dagger^*$Anu Jagannath, $\dagger$Jithin Jagannath, $^*$Yanzhi Wang, and $^*$Tommaso Melodia\\
  \texttt{$\dagger$Marconi-Rosenblatt AI/ML Innovation Lab, ANDRO, NY, USA}
  \\
  \texttt{$^*$Northeastern University, Boston, MA, USA}
}

\begin{document} 
\maketitle

\begin{abstract}
Automatic RF modulation recognition is a primary signal intelligence (SIGINT) technique that serves as a physical layer authentication enabler and automated signal processing scheme for the beyond 5G and military networks. Most existing works rely on adopting deep neural network architectures to enable RF modulation recognition. The application of deep compression for the wireless domain, especially automatic RF modulation classification, is still in its infancy. Lightweight neural networks are key to sustain edge computation capability on resource-constrained platforms. In this letter, we provide an in-depth view of the state-of-the-art deep compression and acceleration techniques with an emphasis on edge deployment for beyond 5G networks. Finally, we present an extensive analysis of the representative acceleration approaches as a case study on automatic radar modulation classification and evaluate them in terms of the computational metrics. 
\end{abstract}


\section{INTRODUCTION}
\label{sec:intro}  

Neural networks are gaining popularity in the radio frequency (RF) applications. Recently, neural networks have been employed to solve numerous challenging RF applications such as RF fingerprinting, blind modulation classification, wireless standard classification, direction of arrival (DoA) estimation, channel equalization, symbol detection, among others\cite{JagannathAdHoc2019,Jagannath18ICCANN,AJagannath21ICC,Ajagannath2022ComST2022,lightAMC,AjagannathCOMNET21}. However, a notable trend is in designing wider and deeper neural networks. Among neural networks, the most popular are convolutional neural networks (CNNs) such as visual geometry group (VGG), AlexNet, GoogleNet, ResNet, DenseNet, among others which although designed for computer vision (CV) applications were shown to perform well for the RF applications \cite{AlexGoogleNet_AMC,exposing,sankhe,Ramjee2019FastDL}. However, unlike CV applications, the \textit{deployability} of the model gains even more priority in the RF world. This can be attributed to the low-end radio platforms with strict computational, memory, and power constraints where the computational resources must host the neural networks in addition to the complex RF transceiver chains and similar intensive signal processing. The deployability of the model is a less investigated topic in the \textit{neural networks for RF} realm except for a few works in this direction \cite{rfEdge,Ajagannath2022MTL,lightAMC}. Deployability must carefully consider the number of layers, neurons, kernel sizes, etc., which significantly impacts the floating point operation (FLOPs), latency, and memory requirements.  Designing neural networks is no trivial task especially considering the edge deployment factor. There is no direct way to estimate the number of layers, neurons, kernel sizes, number of kernels, etc., and must be determined empirically. This aspect was carefully and elaborately investigated in our previous work \cite{AJagannath21ICC} to arrive at an efficient architecture for a modulation and protocol classification problem. 

Neural network compression techniques are the next step in further reducing the redundancy imbibed in the neural network architectures. We define \textit{redundant neuron or layer} as the network parameter which does not impart any additional feature extraction capability to the architecture and can be removed without degrading the performance. In the literature so far, the neural network compression techniques are broadly categorized into knowledge distillation \cite{KD1,KD2}, network pruning \cite{Cun90optimalbrain,Li_l1norm}, parameter quantization \cite{quant}, low-rank approximation \cite{lowRank}, etc. Among these, low-rank approximation decomposes a tensor  belonging to a trained neural network into smaller dimensions to achieve compression. However, low-rank methods can only decompose tensors one-by-one at each layer and cannot identify redundant parameters in the network. At present, network pruning has gained significant traction whereby the redundant parameters of a trained network are determined and removed iteratively to achieve compression. These redundant parameters could be filters, neurons or channels that are determined by some specific criterion such as $\ell_1$-norm, average percentage of zero activations (APoZ), etc. 

Generally, network pruning can be categorized into structured and unstructured pruning. Unstructured pruning does not preserve the network geometry while removing the least significant parameters. On the contrary, structured pruning does not result in irregular and uneven network connections and would fit well in parallel computation platforms. Structured pruning helps achieve direct computational resource savings on embedded platforms and other hardware based systems. This consequently enables modern central processing units (CPUs) and graphical processing units (GPUs) to exploit the computational savings from structured pruning. Lightweight neural networks helps save the sparse parameters in on-chip memory resulting in significant energy savings by reducing the frequent DRAM accesses. In literature often large-sized networks have provided significant performance in solving challenging tasks while small sized networks can be limited in their learning capability. It is therefore appropriate to first design a reasonable sized network that achieves desirable performance followed by pruning them. 

In literature, there are several works that closely evaluates the compressed network performance on challenging CV problems \cite{Cun90optimalbrain,quant,KD1,KD2,kmeans} while only a few remain for the RF domain \cite{rfEdge,Ajagannath2022MTL}. This is the first work that analyzes the effect of different pruning strategies on CBRS radar waveform classification. We demonstrate the classification of CBRS radar waveforms by training a dense VGG16 network. The saliency of the different kernels within the various convolutional layers for the trained task under various pruning criteria are portrayed prior to pruning the network under diverse pruning strategies to achieve significant compression. Specifically, we analyze the effect of an iterative (Setup A) and greedy (Setup B) pruning strategies with different pruning algorithms such as $\ell_1$-norm \cite{Li_l1norm}, APoZ \cite{Hu2016NetworkAPoZ}, and k-means \cite{Hartigan1979Kmeans,kmeans} on the radar classification with a pre-trained VGG16 network. We compress the model by 27.47\% while achieving 84.79\% with no accuracy loss compared to base model with APoZ Setup A. Similarly, under k-means Setup B, the 9.465\% compressed model achieved a 85.17\% accuracy demonstrating a slightly improved accuracy compared to base model. Most notably, we show that the trained base model can be significantly compressed to 99.74\% to achieve a very lightweight model with only 0.04 Million trainable parameters with 80.2\% accuracy and a speedup of 381.52$\times$.

\section{Structured Deep Compression: Pruning Convolutional Filters}
In this section, we introduce the readers to the various pruning strategies that are investigated in this work. The architectures of interest in this article are CNNs. Consequently, we present a systematic walkthrough of the convolutional filter pruning methods. CNNs superior performance in spatial feature extraction can be attributed to their varying sized kernels arranged in the layers to perform strided spatial convolutions to derive condensed feature maps. These feature maps hold the salient features of the input enabling their classification. However, often times the neural network is over-parameterized and would contain redundant filters (kernels) that are expendable. This redundancy opens the door for saliency estimation and redundancy removal without significant loss in accuracy which essentially compresses the network to a smaller size. For instance, well known deep architectures such as VGG16, ResNet50, DenseNet121, among others possess significant redundancy among the different filters and feature maps. Smaller, efficient networks with reduced memory footprint and power consumption promotes the use of CNNs in resource constrained edge platforms. The number of pruned filters will correlate directly with the computational speed up and memory footprint reduction as it reduces the number of trainable parameters. 

We favor structured compression over unstructured owing to their ease of implementation on most widely available general purpose hardware. Structured compression involves the removal of structural elements of the CNN such as filter(s) and/or channel(s) that are well supported by various off-the-shelf deep learning libraries.

We present some preliminaries on feature maps from a FLOPs perspective in order to ease the reader into the various pruning strategies. Consider an input feature map $F_i\in \mathbb{R}^{N_i\times H_i\times W_i}$ to a convolutional layer $i$ where $N_i,\; H_i\;, W_i$ are the number of input channels, height, and width of the input feature map, respectively. Let the layer $i$ contain $N_{i+1}$ convolutional kernels of dimensions $h_k\times w_k\times W_i$ corresponding to a total number of learnable parameters of $N_ih_kw_kW_i$. The convolutional layer maps the input feature maps to output tensor $F_{i+1}\in\mathbb{R}^{N_{i+1}\times H_{i+1}\times W_{i+1}}$ which serves as input for the next convolutional layer $i+1$ by the following transformation. 
\begin{equation}
    F_{i+1}(m_p,n_q) = \sum_{p=1}^{h_k}\sum_{q=1}^{w_k}\sum_{r=1}^{W_i}K_i(p,q,r)F_i(m,n)
\end{equation}
where the spatial location of the output are $m_p=m-p+1$ and $n_q=n-q+1$ considering a unit stride without zero-padding.
In other words, each convolutional kernel in layer $i$ of size $h_k\times w_k\times W_i$ generates one feature map. The total number of FLOPs of layer $i$ is $N_{i+1}H_{i+1}W_{i+1}h_kw_kW_i$. Pruning one filter from the layer $i$ will therefore reduce $H_{i+1}W_{i+1}h_kw_kW_i$ FLOPs and $h_kw_kW_i$ trainable parameters. In a nutshell, pruning or removal of a filter will remove a feature map from the layer output which consequently eliminates a channel from the subsequent convolutional layer.

In a broad sense, we can categorize the structured pruning to the following steps:
\begin{enumerate}
    \item Train a baseline neural network model $\mathcal{F}$ on the target classes.
    \item Rank the filters of layer $l$ as per some metric that determines saliency.
    \item Prune the filter(s) with the least importance to achieve a target pruning rate in layer $l$.
    \item Retrain and fine-tune the pruned network $f$ to achieve the desired accuracy. Repeat step 3 for the target layers until desired compression ratio is achieved.
\end{enumerate}
\subsection{$\ell_1$-norm }
\label{sec:l1}
The authors of \cite{Li_l1norm} proposed an $\ell_1$-norm approach to prune the filters of a layer. According to this approach, the $\ell_1$-norm of the filters are used as a saliency determination metric to physically prune them from the network. For a layer $l$, the $\ell_1$-norm of a filter at index $m$, $K^{l,m}$, can be obtained by computing the sum of its absolute weights, as
\begin{equation}
    \vert \vert K^{l,m}\vert \vert_1  = \sum \vert  k_{i,j} \vert
\end{equation}
Recall here that the number of input channels is the same across filters of a layer. In that sense, the $\ell_1$-norm also represents the average magnitude of its kernel weights. Hence, this gives an estimate of the magnitude of the output feature maps. The filters are ranked or sorted based on their $\ell_1$-norm values and the filters with the smallest values are removed from the layer to achieve desired pruning percentage. 

We note here that determining the importance of the filters based on their numerical values would be an insufficient benchmark. For example, consider two $3\times3$ filters $A = [0.01, 0.005, 0.01;0.01, 0.8, 0.01; 0.03, 0.001, 0.002]$ and $B = [0.5, 0.5,0.5;0.5,0.5,0.5;0.5, 0.5, 0.5]$. The $\ell_1$-norms of these would be 0.8 and 1.5 respectively, causing A to be ranked smaller than B. However, intuitively it can be seen that the filter A places more weightage to the center grid unlike B which considers all the grids equally. Hence, different filters activates different spatial locations of the input features. Consequently, considering only the filter coefficients might prove insufficient. Therefore, it becomes necessary to also consider the feature map activations.
\subsection{k-means}
In this section, we will analyze the effect of clustering on the achievable compression and efficacy in preserving the accuracy. Here, we adopt k-means CNN filter pruning proposed in \cite{k-means} whereby the filters of a layer are subject to k-means clustering. The intuition behind this pruning method is to identify the filters that are located far away from the centroid of the clusters and keep only the ones closest to the center. The outlier filters along with their feature maps and channels of the subsequent convolutional layer will be subject to pruning. 
\subsection{Zero activation analysis}
The two methods discussed above determined the significance of filters based on their coefficients. Recall the numerical example in section \ref{sec:l1}, considering the filter coefficients alone may not correspond to their activations. Therefore, here we will analyze the filter activation-based saliency determination to identify prunable filter(s). The approach in \cite{Hu2016NetworkTA} proposes to consider the average percentage of zero (APoZ) activations of the filter for a sample of the examples after the ReLU mapping. The intuition here is that if a filter is rarely activated, then they do not contribute much to the feature extraction. The APoZ metric is computed for each filter $f$ in a layer $l$ over $M$ examples of the validation set as follows,
\begin{equation}
    APoZ^{f,l,c} = \frac{\sum_{i=0}^{M}\sum_{j=0}^{N} \mathfrak{F}\big(a^{f,l,c}_{ij}==0\big)}{MN}
    \label{eq:apoz}
\end{equation}
where $a^{f,l,c}$ is the activation map of the filter $f$ in layer $l$, $M$ is the the number of examples in the batch, $N$ is the dimension of the filter along channel $c$, and  $\mathfrak{F}\big(bool\big)$ is 1 if $bool$ condition is true and 0 otherwise. The filters with high mean APoZ are subject to pruning to achieve the desired compression ratio.
\section{Case Study: CBRS Radar Waveform Classification}
A 150 MHz bandwidth of the 3.5 GHz Citizens Broadband Radio Services (CBRS) spectrum has been commissioned for shared use by Federal incumbent access users and commercial users. This band was exclusively used by the US Federal government for Navy radar and aircraft communication. Dynamic spectrum access strategies which involve detection of the incumbents in order to free up the spectrum by the secondary commercial users are therefore inevitable to promote a harmonious coexistence between the two entities. This capability also referred to in literature as environmental sensing capability (ESC) currently involves a network of sensors which samples the spectrum to detect the presence of incumbent radar signals. The National Telecommunications and Information Administration (NTIA) has published in \cite{ntia} radar signal parameter bounds to facilitate a device's ESC performance. The literature from the past few years has only looked at radar detection in the CBRS band \cite{cbrsDet1,cbrsDet2,cbrsDet3} which identifies whether a radar signal is present or not (binary classification). However, future ESC systems would require radar identification for comprehensive dynamic spectrum access such that the spectrum reallocation can be performed in a more informed manner. Therefore, in this work for the first time, to the best of the authors' knowledge, we investigate the CBRS radar identification problem which is a multi-label classification. We exploit the CBRS radar waveforms dataset in \cite{nist-cbrs} released by National Institute of Standards and Technology (NIST) which follows the radar signal parameter bounds for ESC compliance testing released by NTIA.

The NIST-CBRS dataset is comprised of five radar modulations: P0N\#1, P0N\#2, Q3N\#1, Q3N\#2, and Q3N\#3 each different from each other in terms of the pulse width, pulses per burst, chirp width (as applicable), and pulse repetition rate as mentioned in \cite{ntia}. The dataset in addition also contains noise examples which corresponds to the scenario where there are no active radar emissions in the spectrum. Hence, the class labels in the radar identification problem are: P0N\#1, P0N\#2, Q3N\#1, Q3N\#2, Q3N\#3, and Noise.

\begin{figure}[ht!]
  \centering
    \includegraphics[width=0.4\textwidth]{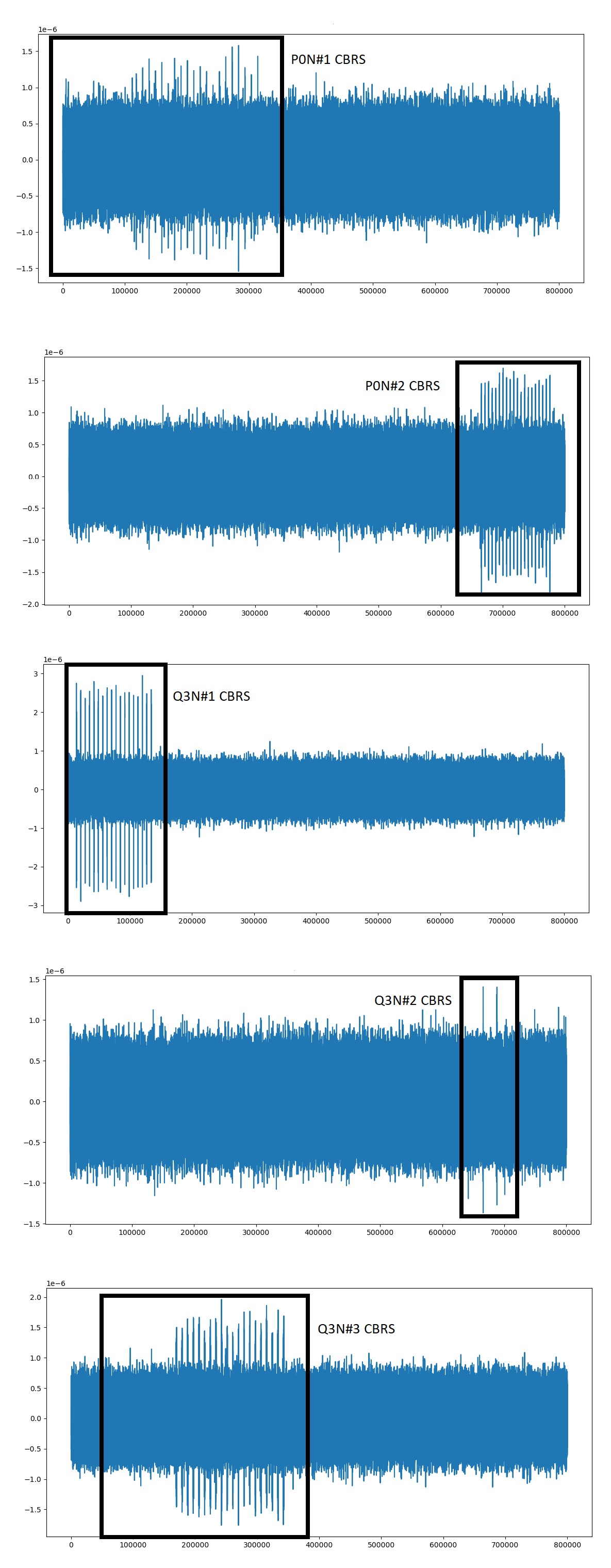}
    \caption{CBRS Radar Waveforms - Time Domain representation at SNR=20dB}
    \label{fig:td}
  \end{figure}
The dataset holds a total of 459 GB worth of samples. We have empirically determined that the distribution of each radar waveform in the dataset is uneven. This stems from the original intention of this dataset - radar detection in the CBRS band. However, to aid the convergence and learning of the neural networks for the radar identification task, for our application we selected same number of examples for each class label to prevent any bias during training. The whole dataset is split into training, validation, and test sets with 4080, 1800 , and 2400 examples respectively. Each example in the dataset contains 800k complex inphase-quadrature (IQ) samples. An example snapshot of a 800k samples time-domain view of all the CBRS radar waveforms of signal-to-noise-ratio (SNR) 20 dB is shown in Figure \ref{fig:td}. Here, we have marked the radar pulses to show the active radar emission. For training the neural networks we mapped the 800k IQ samples into the corresponding time-frequency (TF) domain of dimensions $128\times128\times3$. The time domain to TF transformation is obtained by taking a 128-point short time Fourier transform (STFT) spectrogram of the 800k IQ samples. Note here that the TF map contains three channels each with the spectrogram magnitude, power spectrum density (PSD), and phase mapping. This three channel TF map serves as the input to the VGG16 architecture. 
\section{Experiments}
\subsection{Saliency Analysis: A magnified look}
In this section, we take a closer look at the chosen architectures in terms of the filter saliency. We define saliency as the significance of the filter or layer or neuron in extracting the features and contributing to the subsequent layers. Therefore, a layer with more number of salient filters will have a low compression potential. We determine the saliency of the filters in terms of the $\ell_1$-norm and APoZ metrics to understand the pruning potential of the chosen neural network architectures with these pruning metrics. We choose VGG16 to evaluate the effects of various pruning approaches on the computational and performance fronts. 

\begin{figure}[h]
\centering
\includegraphics[width=\textwidth]{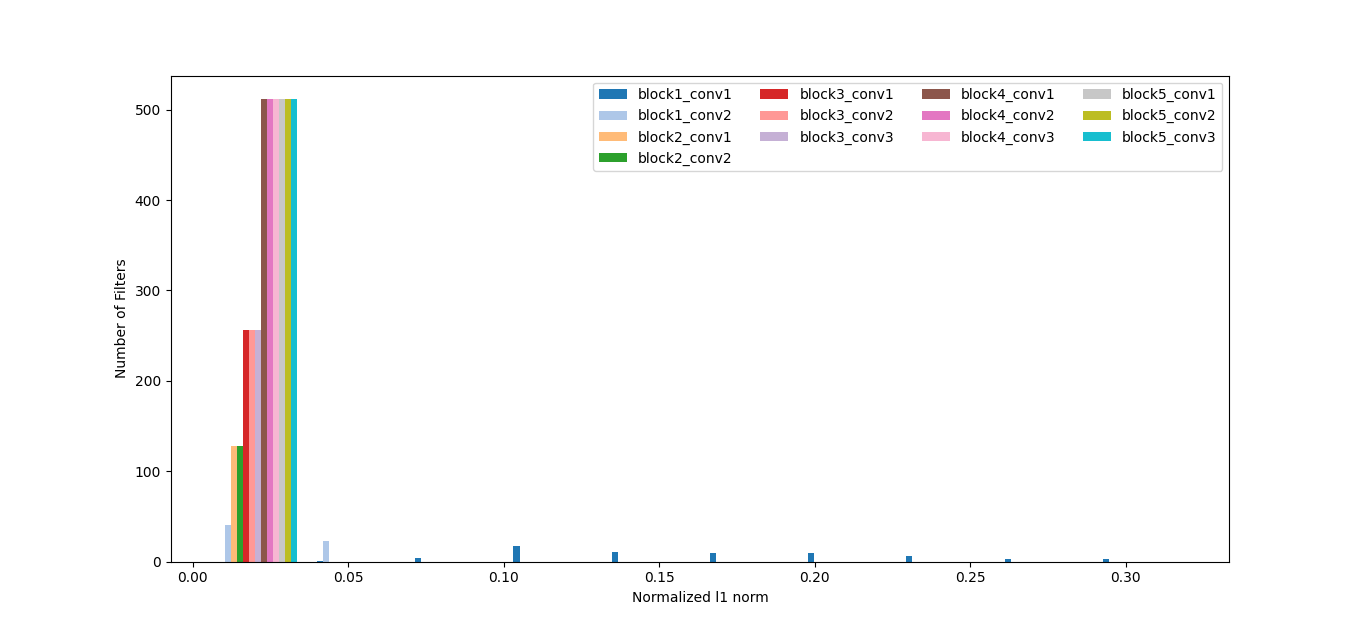}
\caption{$\ell_1$-norm distribution of filters of the VGG16 convolutional layers.}
\label{fig:vl1}
\end{figure}

Figure \ref{fig:vl1} shows the $\ell_1$-norm distribution of all the convolutional layers in the VGG16  model (trained on the NIST CBRS dataset) respectively. As per the $\ell_1$-norm pruning approach, a higher $\ell_1$-norm metric  would indicate highly salient filters and vice-versa. Therefore, if any of the layer has a higher percentage of its filters with a low $\ell_1$-norm value, it would imply that the layer has several redundant and non-salient filters that can be pruned away without much accuracy loss. The Figure \ref{fig:vl1} implies that a majority of the filters in the convolutional layers possess a low $\ell_1$-norm  suggesting a higher pruning potential as per the $\ell_1$-norm pruning approach.

\begin{figure}[h]
\centering
\includegraphics[width=\textwidth]{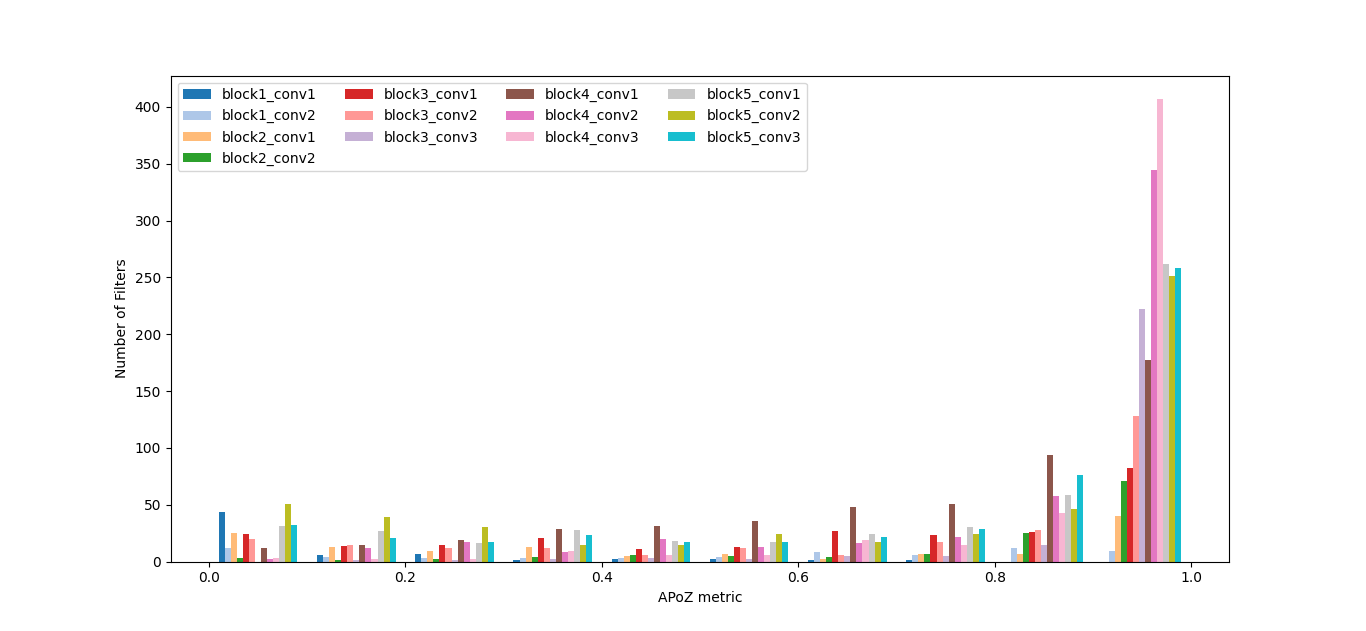}
\caption{APoZ distribution of filters of the VGG16 convolutional layers.}
\label{fig:vapoz}
\end{figure}
On the contrary, the APoZ approach measures saliency in terms of the percentage of zero activations as in Equation \ref{eq:apoz}. Accordingly, a higher APoZ indicates the filter is rarely activated implying less significant feature propagation to the subsequent layer. Hence, a layer with low APoZ for most of its filters will be highly salient and relevant in the feature extraction. The APoZ distribution of all convolutional layers of the VGG16  architecture is depicted in Figure \ref{fig:vapoz}. Figure \ref{fig:vapoz} indicates that the later convolutional layers of the VGG16 architecture has a higher percentage of filters that are rarely activated implying immense compression potential. Next, we will closely evaluate the accuracy and computational tradeoffs incurred by pruning these architectures following different pruning strategies in section \ref{sec:div}.
\subsection{Diverse pruning strategies}
\label{sec:div}
Typically, pruning is performed iteratively one filter at a time and in a layer-by-layer manner. For deep networks, such as VGG16, ResNet50, etc., such an iterative process could take weeks or months of pruning and retraining. The pruning-and-retraining cost in terms of time and computation is one major tradeoff that comes with a target compressed networks. However, iterative pruning process is often suggested and linked to higher classification accuracy. We pose the approaches in two different settings and critically evaluate how the different strategies affect the accuracy as follows,
\begin{enumerate}
    \item Setup A: Candidate filters are selected and iteratively removed from multiple layers simultaneously.
    \item Setup B: Candidate filters are selected and iteratively removed sequentially from the layers on a layer-by-layer basis.
\end{enumerate}
The prune-and-retrain cost as well as the accuracy vary significantly across the different pruning strategies. We will take a closer look at each of these approaches.
\subsubsection{Setup A: Pruning filters iteratively from multiple layers simultaneously}
In this setup, the candidate filters are chosen from each convolutional layer and pruned in an iterative manner from multiple layers at once. The algorithm for Setup A is detailed in Algorithm 1. This approach can be considered a one-shot strategy from a layers perspective while iterative in the number of filters being pruned at once from the target layers. This strategy saves significant prune-and-retrain cost by pruning from all target layers at once. We will evaluate how such a pruned model would compare in terms of accuracy to other strategies.
\begin{algorithm}[H]
\caption{Prune-Retrain Setup A}
\begin{algorithmic}
\State Train a baseline neural network model $\mathcal{F}$
\State Choose pruning strategy $\mathcal{S}$.
\State Determine desired pruning percentage $p$ for each layer. 

\For{\texttt{iter = 1 to MAX\_ITERS}} 
\If{\texttt{iter == 1}}
\State \textbf{Given} model $\mathcal{M} = \mathcal{F}$.
\EndIf
\State Identify prunable filters $\mathbb{F}_l$ $\forall$ target convolutional layers $l\in\mathbb{L} = \{l_1,l_2,\cdots,l_L\}$ of $\mathcal{M}$ based on $\mathcal{S}$ and $p$.
\State Pruning step size $\delta = \min\big(\mathbb{F}_l\big)$
\While{$\mathbb{F}_l \neq \emptyset$}
\State \textbf{Prune:} From all target layers $l$, remove $\delta$ filters from $\mathbb{F}_l$ to obtain pruned model $\mathcal{F}^{iter}$.
\State \textbf{Update} model $\mathcal{M} = \mathcal{F}^{iter}$
\State Recompute filter saliency based on $\mathcal{S}$ and update $\mathbb{F}_l$.
\State \textbf{Retrain} the pruned model $\mathcal{F}^{iter}$ for $N$ epochs.
\EndWhile
\EndFor
\State \textbf{Output:} Compressed and fine-tuned model $\mathcal{F}^*$
\end{algorithmic}
\end{algorithm}

Here, the \texttt{MAX\_ITERS} determine the maximum iterations to achieve the desired pruning percentage for each layer. This will is arbitrarily set such that it is large enough to achieve the desired compression for each layer.

\subsubsection{Setup B: One-shot filter pruning from multiple layers simultaneously}
This approach can be perceived as a greedy approach whereby the candidate prunable filters are removed at once from all the target layers. Unlike, the other two approaches, here the filters do not get to mitigate any feature extraction loss that may arise with the pruning. Rather, here the model is subject to a single-shot pruning and retraining. This pruning strategy denoted as setup B is elaborated in Algorithm 2.
\begin{algorithm}[H]
\caption{Prune-Retrain Setup B}
\begin{algorithmic}
\State Train a baseline neural network model $\mathcal{F}$
\State Choose pruning strategy $\mathcal{S}$.
\State Determine desired pruning percentage $p$ for each layer. 
\State Determine candidate filters $\mathbb{F}_l$ $\forall$ target convolutional layers $l\in\mathbb{L} = \{l_1,l_2,\cdots,l_L\}$ based on $\mathcal{S}$ and $p$.
\State \textbf{Prune:} From all target layers $l$, remove $\mathbb{F}_l$ filters to obtain pruned model $\mathcal{F}^{'}$.
\State \textbf{Retrain} the pruned model $\mathcal{F}^{'}$ for $N$ epochs.

\State \textbf{Output:} Compressed and fine-tuned model $\mathcal{F}^*$
\end{algorithmic}
\end{algorithm}
We would like to point out to the reader that the number of retraining epochs plays a significant role in maintaining the model accuracy. We have empirically determined that greater pruning percentages ($p$) benefit from higher number of retraining epochs. Accordingly, in all of the evaluations we have set the retraining epochs as 10 for $p<50$ and as 30 for $p\geq50$.

\section{Discussion}
We conduct two broad set of experiments (Setup A and Setup B) each with three pruning algorithms - $\ell_1$-norm, APoZ, and k-means - at six pruning ratios yielding a total of thirty six experiments. Having analyzed the saliency distribution, we now look at the pruning potential of the trained architecture. We impose pruning percentage for each convolutional layer to achieve significant model compression. We assess the pruned model in terms of percentage model compression, FLOPs, number of trainable parameters, and model speedup. Below, we define each of these metrics,
\begin{enumerate}
    \item Percentage model compression - ratio of total number of trainable parameters in pruned model to the total number of trainable parameters in the base model.
    \item FLOPs - number of FLOPs in the pruned model.
    \item Trainable parameters - total number of trainable parameters in the pruned model.
    \item Speedup - is a measure of FLOPs reduction and hence the subsequent computational speedup achieved with pruning. It is estimated as the ratio of number of FLOPs in the base model to the number of FLOPs in the pruned model.
\end{enumerate}

\begin{figure}[t]
  \centering
  \begin{minipage}[b]{0.48\textwidth}
    \includegraphics[width=0.99\textwidth]{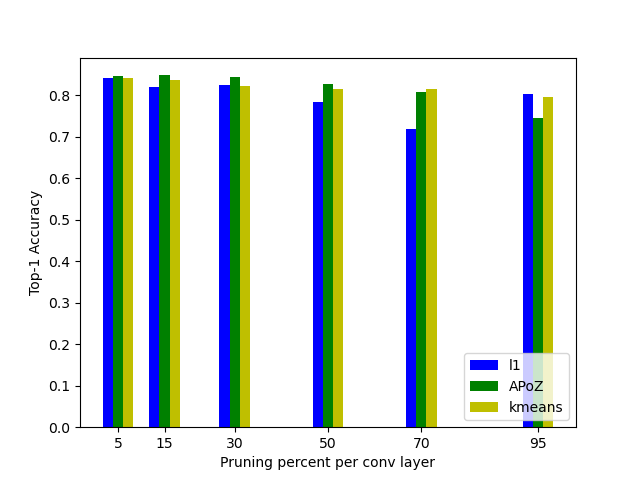}
    \caption{Setup A}
    \label{fig:vA}
  \end{minipage}
  \hspace{1em}
  \begin{minipage}[b]{0.48\textwidth}
    \includegraphics[width=0.99\textwidth]{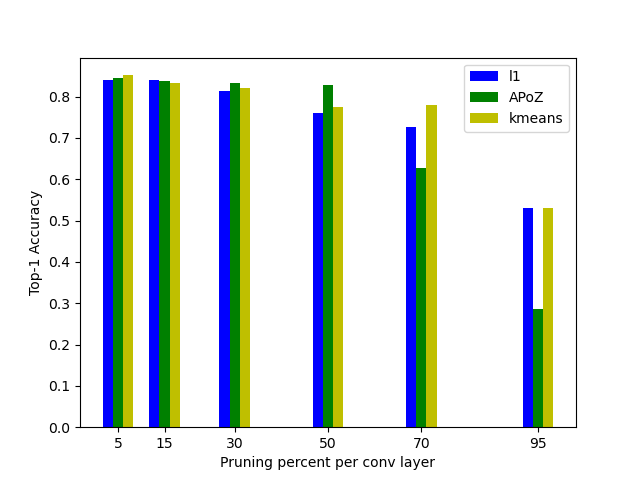}
    \caption{Setup B}
    \label{fig:vB}
  \end{minipage}
\end{figure}
Figure \ref{fig:vA} shows the top-1 accuracy of pruned model at different per layer pruning ratios using the different pruning algorithms. It can be seen that for moderate pruning percentages, APoZ serves as a slightly better pruning algorithm while significant pruning percentages above 50\% for each convolutional layer diminishes its accuracy. Interestingly, at the maximum pruning rates of 95\% per convolutional layer, $\ell_1$-norm yields a slightly higher accuracy. We note here that k-means exhibits stable performance across the various pruning rates under setup A.

Now lets investigate how a greedy strategy (Setup B) will affect the compressed model performance in Figure \ref{fig:vB}. At lower pruning rates ($\leq 30\%$), all the pruning algorithms perform reasonably well under the greedy strategy which makes them ideal to save prune-and-retrain time for lower compression ratios. However, the accuracy takes a steep decline as the pruning rates increase implying for higher pruning rates setup A serves as a good pruning strategy.

We elaborate the in depth comparison of the performance metrics with setup A in Table \ref{tab:setupA}. The 5\% pruning rate from each convolutional layer results in a compression percentage of 9.465. At this low compression rate, the APoZ scheme which identifies and removes filters with zero activations outperform the other two pruning algorithms. This trend remains for all compression rates up to 50\%. Out of these experiments, the most notable one which yields significant speedup and compression is the k-means pruning at 95\% per layer pruning rate. Here, the setup A iterative pruning strategy yields a \textit{very lightweight model that is 99.74\% smaller than the base model with a computational speedup of 381.52$\times$}.
\begin{table}[h!]
    \centering
    \caption{Setup A} \label{tab:setupA}
    \begin{tcolorbox}[tab2,tabularx={|p{2.5 cm}||p{1.2 cm}||p{2.0 cm}||p{2.0 cm}||p{2.0 cm}||p{2.0 cm}||p{2.4 cm}}]
      \textbf{Approaches} & \textbf{Layer \newline pruning $\%$}  & \textbf{Model \newline compression $\%$} &\textbf{FLOPs (Million)}&\textbf{Trainable\newline Parameters (Million)}&\textbf{Speedup}&\textbf{Top-1 \newline Acc. $\%$}\\ \hline\hline
      Baseline & 0 & 0 & \textbf{29.32} & \textbf{140.8} & - & \textbf{84.6}\\ \hline
      \multirow{6}{*}{$\ell_1$-norm} & 5  & 9.465 &26.65 &13.33 &1.1$\times$ &84.16 \\
      & 15  & 27.47 &21.35 &10.68 &1.38$\times$ &82.04 \\ 
       & 30  & 50.78 &14.49 &7.24 &2.03$\times$ &82.38 \\ 
      & 50  & 74.98 &7.36 &3.68 &3.99$\times$ &78.33 \\
      & 70  & 90.93 &2.67 &1.34 &11.03$\times$ &71.92 \\ 
      & 95  & \textbf{99.74} &\textbf{0.077} &\textbf{0.04} &\textbf{381.52$\times$} &\textbf{80.2} \\ \hline
      \multirow{6}{*}{APoZ} & 5  & 9.465 &26.65 &13.33 &1.1$\times$ &\textbf{84.63} \\
      & 15  & 27.47 &21.35 &10.68 &1.38$\times$ &\textbf{84.79} \\ 
      & 30  & 50.78 &14.49 &7.24 &2.03$\times$ &\textbf{84.42} \\ 
      & 50  & 74.98 &7.36 &3.68 &3.99$\times$ &\textbf{82.79} \\ 
      & 70  & 90.93 &2.67 &1.34 &11.03$\times$ &80.79 \\ 
      & 95  & 99.74 &0.077 &0.04 &381.52$\times$ &74.58 \\ \hline
      \multirow{6}{*}{k-means} & 5  & 9.465 &26.65 &13.33 &1.1$\times$ &84.16 \\ 
      & 15  & 27.47 &21.35 &10.68 &1.38$\times$ &83.75 \\ 
       & 30  & 50.78 &14.49 &7.24 &2.03$\times$ &82.29 \\ 
       & 50  & 74.98 &7.36 &3.68 &3.99$\times$ &81.54 \\ 
       & 70  & \textbf{90.93} &\textbf{2.67} &\textbf{1.34} &\textbf{11.03$\times$} &\textbf{81.38} \\
      & 95  & 99.74 &0.077 &0.04 &381.52$\times$ &79.46\\ \hline
      
    \end{tcolorbox}{}
\end{table}


\begin{table}[h!]
    \centering
    \caption{Setup B} \label{tab:setupB}
    \begin{tcolorbox}[tab2,tabularx={|p{2.5 cm}||p{1.2 cm}||p{2.0 cm}||p{2.0 cm}||p{2.0 cm}||p{2.0 cm}||p{2.4 cm}}]
      \textbf{Approaches} & \textbf{Layer \newline pruning $\%$}  & \textbf{Model \newline compression $\%$} &\textbf{FLOPs (Million)}&\textbf{Trainable\newline Parameters (Million)}&\textbf{Speedup}&\textbf{Top-1 \newline Acc. $\%$}\\ \hline\hline
      Baseline & 0 & 0 & \textbf{29.32} & \textbf{140.8} & - & \textbf{84.6}\\ \hline
      \multirow{6}{*}{$\ell_1$-norm} & 5  & 9.465 &26.65 &13.33 &1.1$\times$ &84.08\\ 
       & 15  & \textbf{27.47} &\textbf{21.35} &\textbf{10.68} &\textbf{1.38$\times$} &\textbf{84} \\ 
       & 30  & 50.78 &14.49 &7.24 &2.03$\times$ &81.38 \\ 
       & 50  & 74.98 &7.36 &3.68 &3.99$\times$ &76.04 \\ 
       & 70  & 90.93 &2.67 &1.34 &11.03$\times$ &72.71 \\ 
       & 95  & 99.74 &0.077 &0.04 &381.52$\times$ &52.92 \\ \hline
      \multirow{6}{*}{APoZ} & 5  & 9.465 &26.65 &13.33 &1.1$\times$ &84.42 \\
       & 15  & 27.47 &21.35 &10.68 &1.38$\times$ &83.67 \\ 
       & 30  & 50.78 &14.49 &7.24 &2.03$\times$ &\textbf{83.17} \\ 
       & 50  & 74.98 &7.36 &3.68 &3.99$\times$ &\textbf{82.79} \\ 
       & 70  & 90.93 &2.67 &1.34 &11.03$\times$ &62.75 \\ 
       & 95  & 99.74 &0.077 &0.04 &381.52$\times$ &28.5 \\ \hline
      \multirow{6}{*}{k-means} & 5  & \textbf{9.465} &\textbf{26.65} &\textbf{13.33} &\textbf{1.1$\times$} &\textbf{85.17} \\ 
       & 15  & 27.47 &21.35 &10.68 &1.38$\times$ &83.38 \\ 
       & 30  & 50.78 &14.49 &7.24 &2.03$\times$ &81.96 \\ 
      & 50  & 74.98 &7.36 &3.68 &3.99$\times$ &77.58 \\
       & 70  & 90.93 &2.67 &1.34 &11.03$\times$ &\textbf{78.04} \\ 
       & 95  & 99.74 &0.077 &0.04 &381.52$\times$ &\textbf{53.13} \\ \hline
      
    \end{tcolorbox}{}
\end{table}

Similarly, Table \ref{tab:setupB} demonstrates the performance metrics of the compressed model under different pruning rates and approaches. It is evident here that such a greedy strategy will only suit lower compression rates with the peak performance at 5\% per layer pruning which yields a 9.465\% model compression and a 1.1$\times$ speedup using k-means. Similarly, $\ell_1$-norm achieves an 84\% accuracy for a 27.47\% model compression and speedup of 1.38$\times$. We note here that beyond 50\% pruning rate for each convolutional layer, the greedy setup B is not a suitable pruning strategy. The overall diminished performance at higher pruning rates with setup B in contrast to setup A can be attributed to the lack of iterative prune-and-retrain which facilitates the model to relearn and improve its generalization capability. Intuitively, the advantage of setup B is faster pruning since it does not involve iterative step-by-step retraining. 
\section{Conclusion and Future Work}
In this paper, we investigated diverse pruning strategies and their effect on the classification accuracy with an application on CBRS radar waveform classification. This is the first work that presents an elaborate case study of the saliency analysis and different compression approaches in terms of effective model compression percentage, model speedup, number of FLOPs and trainable parameters, and classification accuracy. We demonstrate that the setup A iterative pruning strategy can yield a significantly compressed model by 99.74\% with a computational speedup of 381.52$\times$ and only 0.04M trainable parameters preserving an accuracy of 80.2\%. We demonstrate that the greedy and faster pruning setup B will serve as a good alternative for lower compression rates. As part of our future work, we plan to analyze more dense architectures and incorporate additional pruning algorithms to test the bounds of compression. We note here that reducing the pruning and retraining time to achieve faster compression is still an open challenge requiring significant investigation.
\appendix    
\bibliography{report} 
\bibliographystyle{spiebib} 

\end{document}